\documentclass[letterpaper, 10 pt, conference]{ieeeconf}  

\IEEEoverridecommandlockouts                              

\usepackage{algorithm}
\usepackage{algpseudocode}
\usepackage{booktabs}
\usepackage{siunitx}
\usepackage[table]{xcolor}
\usepackage{colortbl}
\usepackage{adjustbox}
\usepackage{bm}
\usepackage{tikz}
\usepackage{amsfonts,amsmath,amssymb}
\usepackage{authblk}

\usepackage{nicefrac}
\usepackage{microtype}
\usepackage{todonotes}

\usepackage{graphicx}
\usepackage{booktabs}
\usepackage{tabularx}
\usepackage{array}
\usepackage{multirow}
\usepackage{stfloats}
\usepackage{adjustbox}
\usepackage{float}         
\usepackage[hidelinks]{hyperref}      

\usepackage{xcolor}        
\usepackage{pifont}        

\newcommand{\first}{\tikz\draw[yellow,fill=yellow] (0,0) circle (0.7ex); \ }
\newcommand{\second}{\tikz\draw[lightgray,fill=lightgray] (0,0) circle (0.7ex); }

\newcommand{\alignmark}{\ \tikz\draw[white,fill=white] (0,0) circle (0.7ex);}

\title{\LARGE \bf
InstantSfM: Towards GPU-Native SfM for the Deep Learning Era
}

\author[1,3]{Jiankun Zhong\textsuperscript{*}\thanks{*These authors contributed equally to this work.}}
\author[2]{Zitong Zhan\textsuperscript{*}}
\author[1]{Quankai Gao\textsuperscript{*}}
\author[1]{Ziyu Chen}
\author[1]{Haozhe Lou}
\author[1]{\\Jiageng Mao}
\author[1]{Ulrich Neumann}
\author[2]{Chen Wang}
\author[1]{Yue Wang}
\affil[]{%
    University of Southern California \quad
    \textsuperscript{2}University at Buffalo \quad
    \textsuperscript{3}Tsinghua University%
}

\begin{document}

\maketitle
\thispagestyle{empty}
\pagestyle{empty}
\begin{figure*}[th!]
    \centering
    \begin{minipage}[t]{1.0\linewidth}
        \centering
        \includegraphics[width=\linewidth]{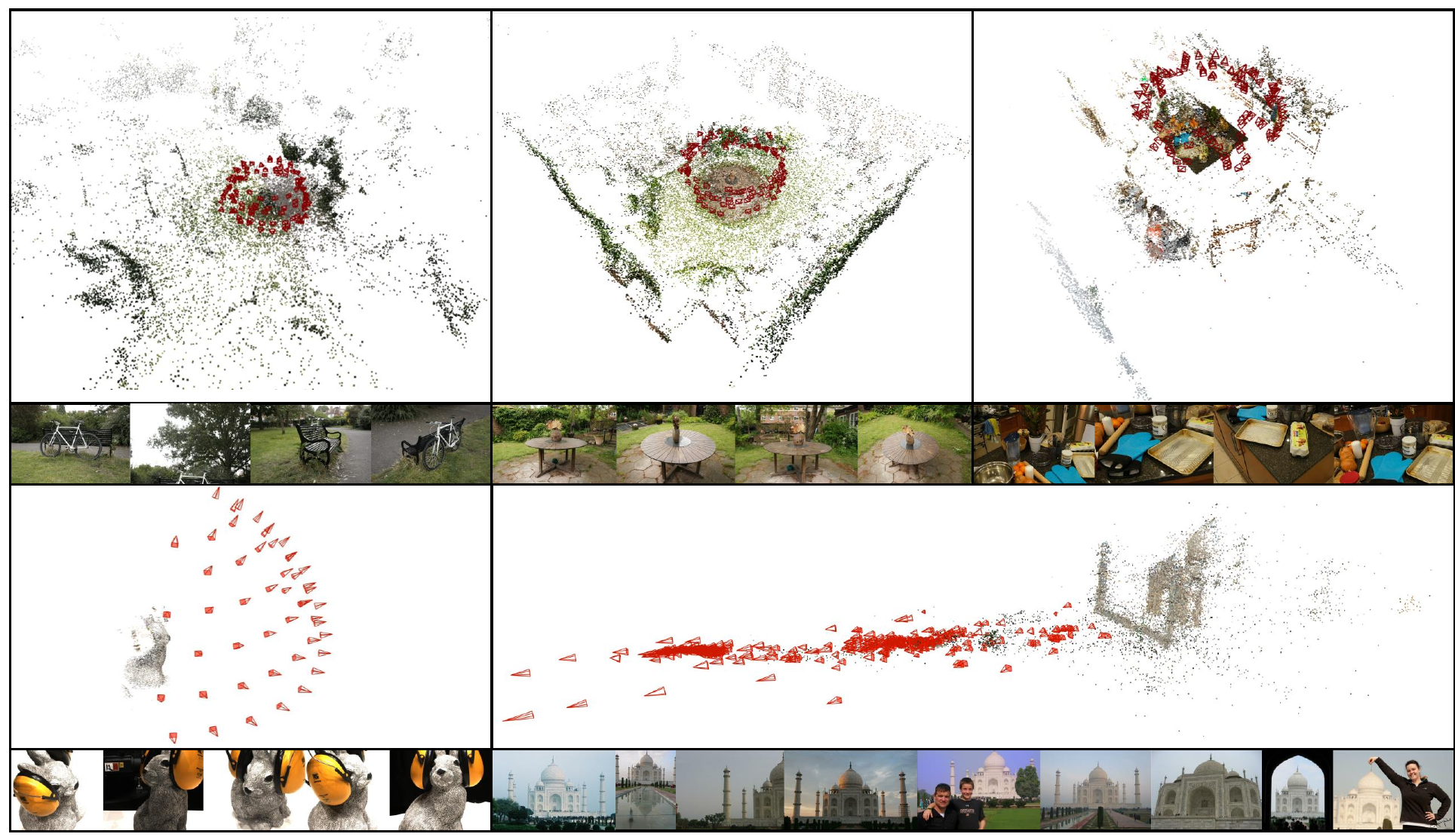} 
    \end{minipage}

    \vspace{-3mm}
    \caption{
      Qualitative results on various datasets. Camera frustums as red are visualized for illustrating the estimated camera poses. Input images for each scene are sampled below each reconstruction.
    }
    \label{fig:qualitative}
    \vspace{-5mm}
\end{figure*}

\begin{abstract}

Structure-from-Motion (SfM) is a fundamental technique for recovering camera poses and scene structure from multi-view imagery, serving as a critical upstream component for applications ranging from 3D reconstruction to modern neural scene representations such as 3D Gaussian Splatting.
However, mature SfM systems remain largely CPU-centric and are usually implemented around traditional optimization toolchains, creating a mismatch with GPU-based, learning-driven vision pipelines.
Recent GPU bundle adjustment (BA) solvers show that sparse second-order optimization can be accelerated substantially, but a complete global SfM system requires additional capabilities beyond faster BA: metric-scale constraints must be incorporated before and during refinement, and the optimizer must remain stable when outlier filtering changes the active residual and parameter sets.
We present InstantSfM, a GPU-native and PyTorch-compatible global SfM system designed to bridge this gap.
InstantSfM embeds metric depth priors into both global positioning and BA through a depth-constrained sparse Jacobian construction, allowing metric information from RGB-D sensors or learned depth models to influence the reconstruction rather than only post-hoc alignment.
It also performs dynamic parameter extraction inside the Levenberg-Marquardt loop, removing currently under-constrained cameras or points from the linear solve while preserving the mapping back to the full reconstruction.
Extensive experiments on diverse datasets demonstrate that InstantSfM achieves state-of-the-art efficiency while maintaining reconstruction accuracy comparable to established classical pipelines and recent learning-based methods, showing up to ${\sim40\times}$ speedup over COLMAP on large-scale scenes.\footnote{Code available at \href{https://github.com/zitongzhan/InstantSfM}{\texttt{github.com/zitongzhan/InstantSfM}}}

\end{abstract}

\section{Introduction}

Structure-from-Motion (SfM) is a foundational technique in computer vision and robotics that jointly estimates camera poses and reconstructs 3D structure from multi-view imagery~\cite{agarwal2011building}. It supports a wide range of applications including autonomous driving~\cite{chen2025omnire} and dynamic scene understanding~\cite{li2024megasam}. SfM also serves as a critical component for downstream 3D vision pipelines, such as multi-view stereo (MVS) dense reconstruction~\cite{hartley1992stereo} and neural scene representations, such as neural radiance fields (NeRF)~\cite{mildenhall2021nerf} and 3D Gaussian Splatting (3DGS)~\cite{kerbl20233d}.

Despite its importance, most existing SfM systems~\cite{schonberger2016sfm,pan2024global} remain largely CPU-based and are implemented in C++-based optimization toolchains~\cite{ceres}. Their architectures reflect design principles established before GPU-native deep learning frameworks became the default environment for vision systems. In contrast, learning-based SfM front-ends~\cite{li2024megasam,wang2025vggt}, learned depth estimators, and neural rendering systems~\cite{ye2025gsplat} are built around parallel GPU computation and are tightly coupled with PyTorch. This architectural gap reduces the practicality of SfM in modern pipelines: reconstruction is often treated as an offline stage outside the learning stack and can require hours to days to process large image collections~\cite{pan2024global}. More broadly, recent robotics research has begun exploring tighter integration between learned models and optimization-based decision modules within unified frameworks~\cite{wang2025imperative,zhan2024imatching}.

This gap has motivated growing interest in GPU-based bundle adjustment (BA)~\cite{zhan2026bundle}, which refines camera poses and 3D structure by minimizing reprojection error. Recent studies demonstrate that large-scale BA can be accelerated on GPUs using PyTorch-native sparse data structures and GPU-parallelized Levenberg-Marquardt (LM) operators~\cite{zhan2026bundle}. However, a complete global SfM pipeline requires more than a fast BA kernel. It must initialize camera positions, handle metric scale, repeatedly filter unreliable structure, and keep the resulting sparse optimization problem numerically stable as the active observations change. 

Two practical challenges are especially important. First, existing GPU-based BA frameworks~\cite{theseus,huang2021deeplm,daba} are primarily designed around reprojection residuals. Robotics and embodied-vision applications often require metric consistency and fusion of additional sensing modalities, such as RGB-D measurements or learned metric depth. Incorporating such depth information into global SfM is not simply a post-processing alignment problem: depth should constrain both the global positioning stage and the final BA, while invalid or partially observed depth values must be handled without destroying the sparse structure used by the solver.

Second, common outliers in feature detection and matching can severely compromise global optimization. Robust loss functions reduce the influence of large residuals, but they do not by themselves remove variables that become geometrically unobservable after visibility or frustum filtering. Conversely, aggressive filtering can leave cameras or 3D points with no active residuals, producing all-zero Jacobian columns, rank-deficient normal equations, and unstable LM updates. This issue is particularly acute in global SfM, where a single unstable global step can corrupt the entire reconstruction.

Motivated by these challenges, we present InstantSfM, a GPU-native and PyTorch-compatible global SfM system. Rather than proposing another isolated BA solver, InstantSfM extends sparse GPU optimization into the full global SfM pipeline, including global positioning, depth-aware refinement, dynamic outlier handling, and reconstruction export.

Specifically, InstantSfM makes two system-level technical contributions. First, we introduce a depth-constrained sparse Jacobian structure that embeds metric depth priors from sensors or monocular depth estimators into both global positioning and BA. This enables metric scale to influence pose and structure estimation during optimization instead of being applied only as a post-hoc alignment. Second, we introduce dynamic parameter extraction for robust global optimization. At each LM iteration, the system identifies the observations and variables that remain geometrically active, compacts them into a solver-ready parameter space on GPU, and scatters the update back to the full reconstruction. This preserves numerical conditioning when outlier filtering changes the effective problem size.

Together, these components yield a scalable and robust global SfM system aligned with modern GPU-based workflows. InstantSfM achieves state-of-the-art efficiency, delivering 1.5$\times$ to 40$\times$ speedup over COLMAP and up to 12$\times$ over GLOMAP across scenes ranging from 100 to 5000 images, while maintaining accuracy comparable to classical SfM pipelines and recent learning-based methods. Beyond speed, InstantSfM enables capabilities that are difficult to express in conventional SfM pipelines, including in-loop metric depth fusion, dynamic optimization over changing active variables, and direct integration with PyTorch-based neural reconstruction workflows.

\section{Related Works}
\label{sec:formatting}

\noindent\textbf{Correspondence Search.}
SfM typically begins with feature detection on a given image set, denoted as $\mathcal{I}=\{I_1, ..., I_n\}$, and extracts either sparse~\cite{lowe2004distinctive,detone2018superpoint} or dense~\cite{sun2021loftr,wang2024efficient} feature descriptors $\mathcal{F}=\{f_1, ..., f_n\}$ to establish view correspondences.
The feature detections are then matched by calculating the feature affinities or more advanced matching algorithms \cite{sarlin2020superglue}. Based on these matching results, homography $\textbf{H}$ and fundamental matrix $\textbf{F}$ are estimated to model pairwise view transformations. Since outliers are inevitably involved in feature extraction, robust estimation algorithms such as RANSAC~\cite{fischler1981random} can greatly improve the accuracy of correspondence search.

\noindent\textbf{Reconstruction.}
Given the estimated scene graph from the correspondence search, reconstruction aims to recover the intrinsic and extrinsic camera parameters, along with the scene structure represented as a set of 3D points triangulated from 2D features~\cite{hartley1997triangulation}. For estimating camera parameters, one prevalent option is incremental reconstruction~\cite{schonberger2016sfm,agarwal2011building,noah2010bundler}, which progressively integrates pairwise view transformations into the existing 3D map reconstruction. Incremental approaches are well-known to be robust against outliers as it iteratively filters outlier and refines pose, but they also suffer from drift and low computational efficiency~\cite{sarlin2022lamar}. Global approaches~\cite{pan2024global,jiang2013global,moulon2013global,rother2003linear} estimate all 3D points and camera poses at once. They employ strategies such as rotation averaging and global positioning, which provide an initial guess of camera poses for BA. Although global SfM approaches may lack per-step outlier filtering or rejection, various designs have been proposed to mitigate drift and error. These methods incorporate 3D cues such as reference planes, line segments, and vanishing points or employ re-triangulation to enhance the robustness of BA~\cite{pan2024global}.

\noindent\textbf{Learning-Based SfM.}
Significant progress has been made in learning-based SfM, including end-to-end differentiable SfM pipelines such as VGGSfM~\cite{wang2024vggsfm}. An orthogonal line of research~\cite{wang2025vggt} approaches the same goal in a feedforward manner by leveraging pretrained image-3D encoder such as DUSt3R~\cite{wang2024dust3r}. However, these methods are either computationally expensive in terms of time and memory, limited to small or moderate-scale scenes, or lack guarantees for generalization to out-of-distribution cases.

\noindent\textbf{Non-Linear Optimization.}
BA is a key optimization method used to refine camera parameters and 3D scene structure by minimizing the reprojection error. It boils down to solving a non-linear least squares problem, whose size scales with the number of images and 3D points. Since there is no general closed-form solution, most of the literature focuses on developing efficient second-order numerical solvers for BA, ranging from Gauss-Newton methods to the more robust Levenberg-Marquardt (LM) algorithm~\cite{ranganathan2004levenberg}. Widely-used BA frameworks are either C++-based, enabling parallelization on CPU such as Ceres~\cite{ceres} and GTSAM~\cite{dellaert2022borglab}, or PyTorch implementations such as DeepLM~\cite{huang2021deeplm}, which utilize PyTorch's autograd engine for Jacobian calculations.

\noindent\textbf{GPU-based BA.}
Recent work has explored GPU-based BA optimizations~\cite{zhan2026bundle}, demonstrating significant acceleration through sparse matrix operations and cuSPARSE implementations. Building upon these sparse-aware techniques, we extend the approach to encompass both BA and global positioning within a unified PyTorch framework.
Although existing tools achieve high accuracy, they either remain inefficient for large-scale BA problems \cite{ceres} or cannot generalize to more types of problems\cite{huang2021deeplm}.

\section{Background of Global SfM} 
\label{sec:Pipeline}
Our contributions center on extending sparsity-aware non-linear least squares optimization to support metric depth constraints and robust outlier handling within a unified GPU framework. We introduce the formulations of global positioning (GP) and BA and analyze the sparse structure of their Jacobians, which directly motivates our technical designs in Sec.~\ref{sec:method}.

\subsection{Global Camera Pose Estimation}
Global camera pose estimation plays an important role in initializing camera extrinsics before iterative BA. It consists of rotation averaging~\cite{hartley2013rotation} and translation averaging~\cite{govindu2001combining,jiang2013global}. GLOMAP~\cite{pan2024global} proposes GP as a more robust alternative to translation averaging by jointly estimating locations of camera centers and 3D points with fixed camera rotations. Denote the number of cameras and points by $C$ and $P$, respectively. The optimization objective can be formulated as follows:
\begin{equation}
    \label{eq:GP_objective}
    \bm{\theta} = \arg\min_{\textbf{X}, \textbf{t}, s}\sum_{i=1}^C\sum^P_{j=1}\rho (||\underbrace{\textbf{v}_{ij}-s_{ij}(\textbf{X}_j-\textbf{t}_i)}_{\mathbf{u}_{ij}}||^2_2),
\end{equation}
where optimizable parameters $\bm{\theta}$ include 3D point locations $\textbf{X}\in\mathbb{R}^{P\times3}$, camera centers $\textbf{t}\in\mathbb{R}^{C\times3}$ and scalar scales $s_{ij}$ for every valid observation; $\rho$ is Huber~\cite{huber1992robust} robustifier and $\textbf{v}$ denotes pixel ray direction vectors from camera centers to 2D feature points. After convergence, each triangulated 3D point $\textbf{X}_k$ should lie in its corresponding pixel ray emitted from visible cameras ensuring global consistency. The per-observation scale variables $s_{ij}$ in this formulation serve as the entry point for our depth-constrained extension in Sec.~\ref{sec:method}.
\subsection{Bundle Adjustment} 
Global camera pose estimation estimates camera rotations, camera centers, and 3D points independently with limited accuracy. BA can further refine the above estimations together with camera intrinsics. Similar to GP, BA is formulated as the least squares problem, minimizing the reprojection errors $\mathbf{r}\in\mathbb{R}^{2\times CP}$ of 3D points across visible camera views: 
\begin{equation}
    \label{eq:BA_objective}
    \bm{\theta} = \arg\min_{\textbf{X}, \bm{\zeta},\textbf{K}}\sum_{i=1}^{C} \sum_{j=1}^{P} \rho(|| \underbrace{\Pi (\bm{\zeta}_i, \mathbf{X}_j, \mathbf{K}_i) - \mathbf{x}_{ij}}_{\mathbf{r}_{ij}} ||_2^2),
\end{equation}
where optimizable parameters $\bm{\theta}$ include camera intrinsics $\textbf{K}\in\mathbb{R}^{C\times 1}$ and extrinsics $\bm{\zeta}=\{\textbf{t},\textbf{o}\}\in\mathbb{R}^{C\times 7}$ consisting of camera centers $\textbf{t}\in\mathbb{R}^{C\times 3}$ and quaternions $\textbf{o}_i\in\mathbb{SO}(3)$. $\mathbf{r}_{ij}$ denotes the reprojection error between a 2D feature point $\mathbf{x}_{ij}$ on the $i$-th camera and the $j$-th 3D point $\mathbf{X}_j$ projected to the same camera. We extend this formulation with metric depth residuals and robust outlier handling in Sec.~\ref{sec:method}.

\subsection{Levenberg-Marquardt Algorithm}
We use the LM algorithm for both the GP and BA objectives. It solves the following linear system for an update step $\Delta\bm{\theta}$ at each iteration,
\begin{equation}
    \label{eq:LM}
(\mathbf{J}^\top \mathbf{J} + \lambda \operatorname{diag}(\mathbf{J}^\top \mathbf{J})) \Delta \boldsymbol{\theta} = -\mathbf{J}^\top \mathbf{r},
\end{equation}
where $\mathbf{J}=\frac{\partial \mathbf{r}}{\partial \bm{\theta}}$ is the Jacobian matrix. For BA, $\mathbf{J}\in\mathbb{R}^{2CP\times (7C+3P)}$, which is infeasible to store densely in GPU memory even at moderate scale. Since the scale of $\mathbf{J}$ directly determines runtime, we build upon sparse-aware BA techniques~\cite{zhan2026bundle} and extend the sparse data structure to encompass global positioning within a unified framework.

\section{Methodology}
\label{sec:method}

InstantSfM introduces two technical components that make GPU-native global SfM practical. First, we propose a \textit{depth-constrained Jacobian structure} (Sec.~\ref{sec:method_depth}) that embeds metric depth priors as live constraints directly inside GP and BA, allowing scale to propagate to cameras and points through the shared sparse structure. Second, to ensure numerical robustness under outlier contamination, we introduce \textit{dynamic parameter extraction} (Sec.~\ref{sec:robust_outlier_removal}), which re-evaluates geometric validity during each LM iteration, compacts the active observations and variables, and removes temporarily under-constrained elements from the linear solve.

\subsection{Depth-Constrained Jacobian Structure}
\label{sec:method_depth}
SfM is scale-ambiguous: camera poses and 3D structure can only be recovered up to an unknown global scale factor from image observations alone. Classical pipelines such as COLMAP have no mechanism to incorporate metric depth available from RGB-D sensors or monocular depth estimators into the reconstruction process itself; depth can only be applied in post-processing to rigidly align the finished reconstruction to a known scale. This fitting is incapable of correcting accumulated drift or resolving local scale inconsistencies that arise during reconstruction, nor does it offer any benefit in estimating the camera poses and 3D structure, leaving the metric depth priors largely unexploited. However, recovering reconstructions at consistent metric scale is important for robotics applications such as physics-based simulation~\cite{lou2024robogsphysicsconsistentspatialtemporal} and cross-modal depth fusion~\cite{xuan2025mrasfmmulticamerareconstructionaggregation}, where absolute distances must be preserved. To fulfill this need, our pipeline instead embeds metric depth priors as joint residual constraints directly inside both the GP and BA optimization stages, allowing depth prior to facilitate reconstruction through the shared sparse Jacobian structure. We first introduce the formulation of depth-constrained optimization and then describe how we implement it in a GPU-friendly manner within the LM optimizer.

\noindent\textbf{GP.} The GP objective \eqref{eq:GP_objective} contains a per-observation scale variable $s_{ij}$, which represents the inverse depth of point $\mathbf{X}_j$ as seen from camera $i$: since $\mathbf{v}_{ij}$ is a unit-normalized ray direction, $s_{ij}$ scales the ray to reach $\mathbf{X}_j$. When a metric depth measurement $\hat{d}_{ij}$ is available from the depth map of camera $i$, we fix this otherwise free variable to its known value $s_{ij} = 1/{\hat{d}_{ij}}$, removing it from the set of optimized parameters. Substituting into~\eqref{eq:GP_objective}, the depth-constrained residual becomes:
\begin{equation}
\mathbf{u}_{ij} = \textbf{v}_{ij} - \frac{\textbf{X}_j - \textbf{t}_i}{\hat{d}_{ij}},
\end{equation}
and the GP objective for points with valid depth measure:
\begin{equation}
\bm{\theta} = \arg\min_{\textbf{X}, \textbf{t}}\sum_{i=1}^C\sum^P_{j=1}\rho (||\textbf{v}_{ij}-\frac{(\textbf{X}_j-\textbf{t}_i)}{\hat{d}_{ij}}||^2_2).
\end{equation}

\noindent\textbf{BA.} To incorporate metric depth into the BA stage, we introduce an additional depth residual term following \cite{huang2025vipe}:
\begin{equation}
\mathbf{r}^d_{ij}=\frac{1}{\text{Depth}(\bm{\zeta}_i, \mathbf{X}_j, \mathbf{K}_i)}-\frac{1}{\hat{d}_{ij}},
\end{equation}
where $\text{Depth}(\bm{\zeta}_i, \mathbf{X}_j, \mathbf{K}_i)$ is the depth calculated from estimated cameras and points, and the whole BA formulation: 
\begin{equation}
    \label{eq:BA_objective_with_depth}
    \bm{\theta} = \arg\min_{\textbf{X}, \bm{\zeta},\textbf{K}}\sum_{i=1}^{C} \sum_{j=1}^{P} \rho(\mathbf{r}_{ij}+\lambda_d\mathbf{r}^d_{ij}),
\end{equation}
where $\lambda_d$ is the weighting parameter for the depth constraint. 
A key challenge when incorporating monocular depth priors is that depth map is inherently incomplete: regions such as sky, specular surfaces, or distant clutter produce invalid estimates. Naively masking these points would remove useful reprojection constraints and destabilize the sparse normal equations. We address this through a \textit{heterogeneous depth-masking} scheme operated directly on the sparse Jacobian. 

\noindent
\textbf{In GP}, the primary role is to provide a metric-scale initialization for 3D point positions. Points with valid depth measurements serve as metric anchors: since their scale $s_{ij}$ is already fixed by the known depth ($s_{ij} = 1/\hat{d}_{ij}$), they are algebraically \emph{pinned} by removing their $s_{ij}$ variables from the Jacobian. These anchored points then propagate metric scale constraints through shared camera centers to points lacking valid depth, which remain as free parameters optimized normally. This selective treatment requires specialized sparse block assembly routines that dynamically partition all point observations into constrained and unconstrained subsets, a design not yet considered by existing GPU BA frameworks.

Traditional CPU-based solvers would handle this by checking validity of each observation sequentially and adding them to the sparse Jacobian dictionary \cite{zhan2026bundle} one-by-one, but this approach is inefficient on GPU due to thread divergence and irregular memory access patterns. Instead, our goal is to design a unified Jacobian structure that can be constructed in a single pass, while still effectively removing the influence of invalid depth measurements on the optimization.

To handle invalid depth entries, we introduce a binary mask $m_{ij}$ indicating the validity of each depth measurement, where $m_{ij}=1$ if $\hat{d}_{ij}$ is valid and $0$ otherwise. This flag partitions the full scale parameter set into pinned and free subsets:
\begin{align}
\mathcal{S}_{\mathrm{pin}} &= \{(i,j)\mid m_{ij}=1\}; \mathcal{S}_{\mathrm{free}} = \{(i,j)\mid m_{ij}=0\}.
\end{align}
Only $\mathcal{S}_{\mathrm{free}}$ contributes columns to the Jacobian $\mathbf{J}$ entering the normal equations; columns of $\mathcal{S}_{\mathrm{pin}}$ are structurally excluded before the LM solve. This column-level exclusion is computed via a parallel set-membership test on GPU that identifies the free-scale indices and compacts the Jacobian accordingly.

Although the pinned scales are removed as optimization variables, their metric information is not lost. It is implicitly transferred to the rest of the scene through the \emph{shared camera-center columns} of $\mathbf{J}$. Specifically, from the GP residual $\mathbf{u}_{ij} = \mathbf{v}_{ij} - s_{ij}(\mathbf{X}_j - \mathbf{t}_i)$, the Jacobian block with respect to the camera center is $\partial \mathbf{u}_{ij}/\partial \mathbf{t}_i = s_{ij}\mathbf{I}$, so every observation from the same camera $i$ contributes to the same camera-center block-column of $\mathbf{J}$. In the normal equations $\mathbf{J}^\top\mathbf{J}$, this yields a non-zero off-diagonal coupling between any free scale $s_{i'j'}\in\mathcal{S}_{\mathrm{free}}$ and the camera center $\mathbf{t}_i$ shared with pinned observations. The pinned observations impose metrically-scaled gradients on $\mathbf{t}_i$ (since their $s_{ij}=1/\hat{d}_{ij}$ encodes the known metric depth), and through this coupling the updated $\mathbf{t}_i$ in turn drives the free-scale update, propagating metric consistency across the entire scene.

\noindent
\textbf{In BA}, we preserve observations without valid depth in the Jacobian, but we prevent them from becoming metric anchors. 
Instead, we assign their reference inverse depth to a small sentinel value, which acts as a weak far-depth prior after weighting while preserving the reprojection constraints.
This unified formulation handles partial depth coverage without removing useful image observations or introducing a separate sparse structure for valid and invalid depth entries. 

To implement this efficiently on GPU, we use the same validity mask $m_{ij}$ to modulate the depth residuals during Jacobian construction. Specifically,
invalid-depth entries are handled by defining the effective inverse-depth reference uniformly for all observations as:
\begin{equation}
\tilde{d}^{-1}_{ij} = m_{ij} \cdot \hat{d}_{ij}^{-1},
\end{equation}
so the depth residual $\mathbf{r}^d_{ij} = 1/\text{Depth}(\cdot) - \tilde{d}^{-1}_{ij}$ no longer uses an invalid measurement as a metric target. Because the formula is structurally identical regardless of validity, the residual tensor is evaluated in one uniform, efficient GPU operation.

\begin{figure}[t]
    \centering
    \begin{minipage}[t]{1.0\linewidth}
        \centering
        \includegraphics[width=\linewidth]{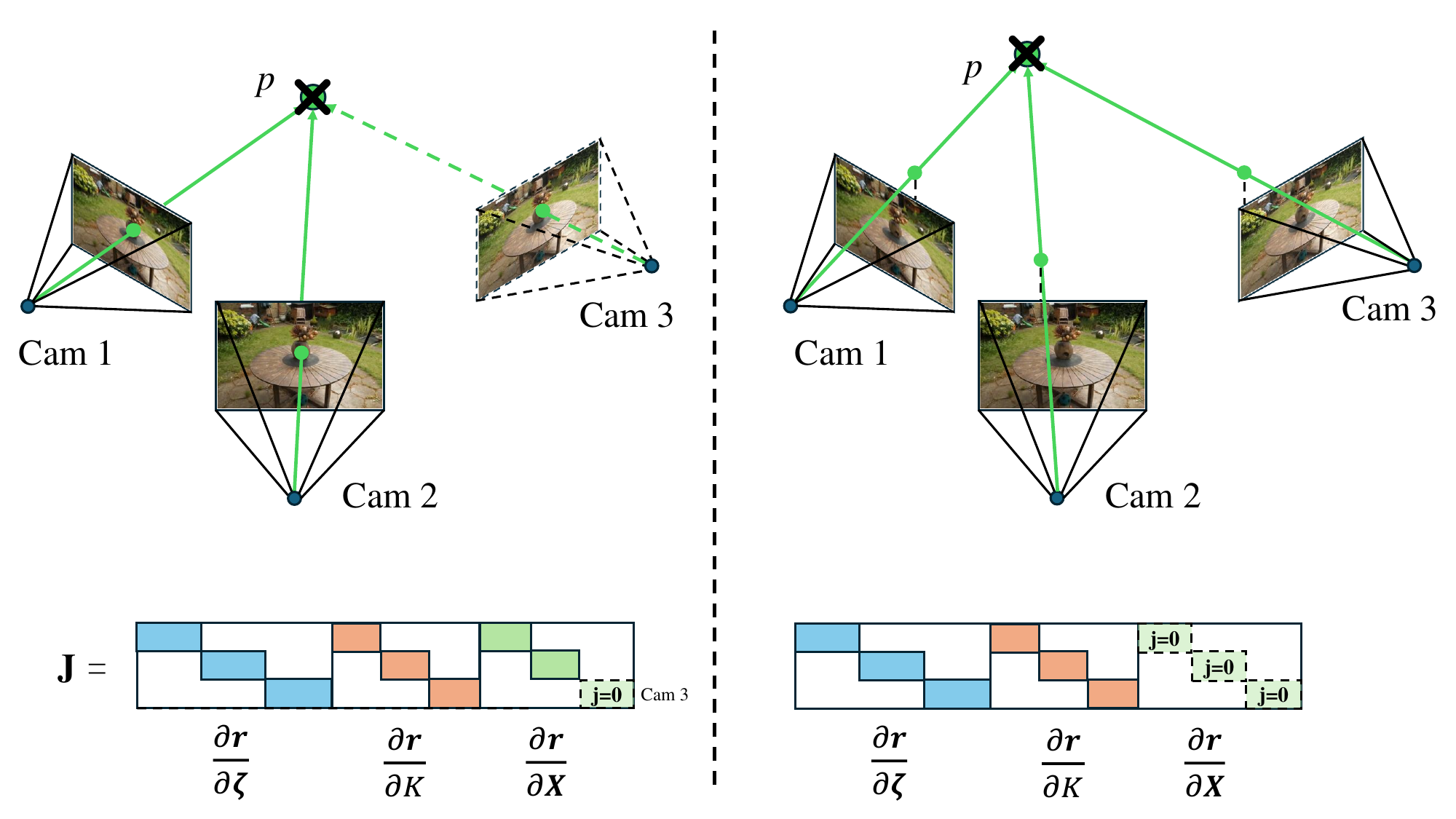} 
    \end{minipage}

    \vspace{-1mm}
    \caption{
       We illustrate the motivation behind our robust outlier removal. Outliers often arise from noisy feature extraction or incorrect feature matching and are typically filtered based on their visibility across multiple cameras. A 3D point is treated as an outlier and removed if it is observed by fewer cameras than a predefined threshold (3 in the figure above) or if none of its reprojections fall within any camera frustum. However, this ``removal" can lead to numerical issues during optimization (see Sec.~\ref{sec:robust_outlier_removal}).
    }
    \vspace{-4mm}
    \label{fig:filter}
\end{figure}
\subsection{Robust Outlier Removal}
\label{sec:robust_outlier_removal}
As illustrated in Fig.~\ref{fig:filter}, outliers are common in a SfM system and can adversely affect the accuracy of LM convergence. These outlier 3D points typically arise when a point is observed by only a small number of cameras, or when its reprojections lie outside the viewing frusta of all cameras.
However, excluding outliers in a global SfM is inherently difficult because point visibility and reprojection behavior change as camera poses and 3D structure are updated during optimization: a point classified as an outlier early may become valid later and vice versa. 

Prior works typically remove such outliers only once during preprocessing or rely on fixed robust loss functions during optimization, due to limitations of common optimization frameworks that do not natively support iterative, geometry-aware pruning inside the LM loop.
This one-shot visibility check risks discarding useful constraints or retaining harmful ones, because the nonlinearity of reprojection geometry tightly couples point validity with the current estimate: small updates to intrinsics or poses can shift reprojections across frustum boundaries or dramatically alter the inlier-outlier status of marginal points. A heuristic-based cost function cannot fully resolve the outlier problem either, because it merely down-weights large residuals without considering geometric visibility or consistency across multiple views. As a result, the optimizer may preserve geometrically invalid points that continue to bias the solution.

A further complication arises when a point temporarily loses all valid observations. One might consider simply zeroing out the residuals and gradients associated with such points, but this creates an ill-conditioned optimization problem as the corresponding columns of the Jacobian matrix become identically zero, producing rank deficiency in the normal equations and leading to singular or near-singular linear systems. In practice, this destabilizes LM updates, amplifies numerical noise, and can completely break convergence.
These factors motivate our outlier rejection that is performed jointly with the evolving optimization state.

Our method integrates outlier detection directly into the optimization through a \textit{dynamic parameter extraction} mechanism that maintains numerical stability while adapting to evolving geometry.
The key distinction from one-shot pruning or robust losses is that we restructure the optimization problem at each iteration to include only geometrically valid observations and their associated active parameters. Specifically,

1) We construct the complete set of potential observations from all registered tracks and cameras at the beginning of each LM iteration. We then perform a geometry validity check; each 3D point is projected into its observing cameras' coordinate frames, and observations where the projected depth is non-positive or out-of-frustum are marked as invalid. This yields a filtered observation set with corresponding camera and point indices:
\begin{equation}
\mathcal{O}_{\text{valid}} = \{(i_c, i_p) \mid z_{i_c,i_p} > 0.1\},
\end{equation}
where $i_c$ and $i_p$ denote the array indices of the full camera and point parameters, respectively.
The critical challenge is that after filtering, many cameras and points may have no remaining valid observations, yet their parameters still occupy entries in the full parameter vector $\mathbf{x} \in \mathbb{R}^{n_{\text{total}}}$. Including these inactive parameters in the optimization would yield a Jacobian $\mathbf{J} \in \mathbb{R}^{m \times n_{\text{total}}}$ with all-zero columns, which in turn makes $\mathbf{J}^\top\mathbf{J}$ singular and the LM update unsolvable.

2) Then we identify the unique set of cameras and points that participate in at least one valid observation, 
\begin{align}
\mathcal{C}_{\text{active}} &= \{\text{unique}(i_c \mid (i_c, i_p) \in \mathcal{O}_{\text{valid}})\} \\
\mathcal{P}_{\text{active}} &= \{\text{unique}(i_p \mid (i_c, i_p) \in \mathcal{O}_{\text{valid}})\},
\end{align}
where the unique set operation is achieved in a highly efficient SIMD manner, $e.g.$ by \texttt{torch.unique}.
We then extract only these active parameters into a compact representation:
\begin{align}
\hat{\mathbf{c}} &= \mathbf{c}[\mathcal{C}_{\text{active}}] \in \mathbb{R}^{|\mathcal{C}_{\text{active}}| \times d_c} \\
\hat{\mathbf{p}} &= \mathbf{p}[\mathcal{P}_{\text{active}}] \in \mathbb{R}^{|\mathcal{P}_{\text{active}}| \times 3}
\end{align}
where $\mathbf{c}$ and $\mathbf{p}$ are the full camera and point parameter arrays, and $d_c$ is the camera parameter dimension. Crucially, we also construct a remapping from the original observation indices to the compacted index space:
\begin{equation}
\tilde{i}_c = \text{rank}(i_c \text{ in } \mathcal{C}_{\text{active}}), \quad \tilde{i}_p = \text{rank}(i_p \text{ in } \mathcal{P}_{\text{active}})
\end{equation}
This remapping ensures that the residual computation in the forward pass correctly indexes into $\hat{\mathbf{c}}$ and $\hat{\mathbf{p}}$.

The resulting optimization problem operates entirely in the compacted space with $\hat{n} = |\mathcal{C}_{\text{active}}| \cdot d_c + |\mathcal{P}_{\text{active}}| \cdot 3$ parameters and $m = |\mathcal{O}_{\text{valid}}| \cdot 2$ residuals (for 2D reprojection). Critically, every parameter in $\hat{\mathbf{x}} = [\hat{\mathbf{c}}; \hat{\mathbf{p}}]$ appears in at least one residual by construction, guaranteeing that the Jacobian $\hat{\mathbf{J}} \in \mathbb{R}^{m \times \hat{n}}$ has no all-zero columns. The normal equations $\hat{\mathbf{J}}^\top\hat{\mathbf{J}}\,\delta\hat{\mathbf{x}} = -\hat{\mathbf{J}}^\top\mathbf{r}$ are thus full-rank (up to gauge freedom), and the LM step remains well-conditioned even when a large fraction of points are temporarily invalid.

3) After computing the update $\delta\hat{\mathbf{x}}$ via preconditioned conjugate gradient, we scatter the updates back to the original parameter arrays using the inverse of the index mappings. Parameters not in $\mathcal{C}_{\text{active}}$ or $\mathcal{P}_{\text{active}}$ remain unchanged for that iteration. At the next iteration, visibility is re-evaluated with the updated poses and structure, and the compaction process repeats, potentially activating different subsets of parameters as geometry evolves.
This \textit{dynamic parameter extraction} mechanism enables our optimizer to seamlessly handle points that transition between valid and invalid states throughout the optimization. Unlike static outlier rejection schemes that commit to a fixed residual structure, or robust kernels that merely down-weight residuals without addressing rank deficiency, our approach fundamentally adapts the problem dimensionality to match the current geometric configuration. This ensures numerical stability across all optimization stages while naturally excluding degenerate observations.

\section{Experiment}
\label{sec:Experiment}

\begin{table*}[!t]
    \centering
    \caption{Comparison on the MipNeRF360 dataset~\cite{barron2022mip} using the NVS metric.}
    \resizebox{\linewidth}{!}{
    \begin{tabular}{lcccc|cccc| cccc}
        \toprule
        & \multicolumn{4}{c}{\textbf{PSNR} $\uparrow$} & \multicolumn{4}{c}{\textbf{SSIM}$\uparrow$} & \multicolumn{4}{c}{\textbf{LPIPS}$\downarrow$}\\
        \cmidrule(lr){2-5} \cmidrule(lr){6-9} \cmidrule(lr){10-13}
        & COLMAP & GLOMAP & VGGSfM & Ours & COLMAP & GLOMAP & VGGSfM & Ours & COLMAP & GLOMAP & VGGSfM & Ours \\
        \midrule
        bicycle &  20.75 & 25.97 & 25.81 & 25.73 & 0.576 & 0.791 & 0.783 & 0.780 & 0.384 & 0.156 & 0.162 & 0.162 \\
        bonsai & 19.86 & 32.96 & 32.56 &32.06 & 0.619 & 0.957 & 0.953 & 0.947 & 0.490 & 0.054 & 0.056 & 0.061\\
        counter & 29.10 & 29.26 & 29.29 & 29.23 & 0.913 & 0.916 & 0.916 & 0.915 & 0.092 & 0.091 & 0.092 & 0.092 \\
        garden  & 27.91 & 27.80 & 27.74 & 27.66 &  0.878 & 0.876 & 0.875 & 0.869 & 0.072 & 0.073 & 0.073 & 0.077\\
        kitchen  & 32.09 & 16.11 & 15.96 & 27.79 & 0.947 & 0.388 & 0.390 & 0.845 & 0.046 & 0.880 & 0.878 & 0.117 \\
        room  & 31.54 & 31.96 & 31.91  & 31.04 & 0.936 & 0.937 & 0.936 & 0.925 & 0.085 & 0.083 & 0.085 & 0.102\\
        stump  & 27.21 & 26.61 & 26.51 & 25.53  & 0.797 & 0.760 & 0.758 & 0.711 & 0.143 & 0.164 & 0.168 & 0.184\\
        \midrule
        Average & 26.92 & \second 27.24 \alignmark & 27.11 & \first 28.43 \alignmark & 0.809 & \second 0.804 \alignmark & 0.802 & \first 0.856 \alignmark & \second 0.187 \alignmark & 0.214 & 0.216 & \first 0.113 \alignmark \\
        \bottomrule
    \end{tabular}}
    \par\vspace{1mm}
    {\footnotesize\raggedright \first best, \second second best. All experiments use \texttt{images\_4} from the dataset, which is $4\times$ downsampled from the original 4K raw data, resulting in around 1K resolution.\par}
    \label{tab:mipnerf360_nvs_table}
\end{table*}

\begin{table*}[!t]
    \centering
    \caption{Quantitative evaluation for SfM methods on DTU dataset.}
    \label{tab:dtu_rendering_metric}
    \resizebox{\linewidth}{!}{
    \begin{tabular}{lcccc|cccc|cccc}
        \toprule
        & \multicolumn{4}{c}{\textbf{PSNR} $\uparrow$} & \multicolumn{4}{c}{\textbf{SSIM} $\uparrow$} & \multicolumn{4}{c}{\textbf{LPIPS} $\downarrow$} \\
        \cmidrule(lr){2-5} \cmidrule(lr){6-9} \cmidrule(lr){10-13}
        & COLMAP & GLOMAP & VGGSfM & Ours & COLMAP & GLOMAP & VGGSfM & Ours & COLMAP & GLOMAP & VGGSfM & Ours \\
        \midrule
        scan002 & 21.83  & 20.77  & 23.32  & 22.80  & 0.872  & 0.862  & 0.867  & 0.868  & 0.094  & 0.166  & 0.094  & 0.096  \\
        scan016 & 24.94  & 24.18  & 22.26  & 24.04  & 0.892  & 0.891  & 0.869  & 0.899  & 0.075  & 0.237  & 0.085  & 0.067  \\
        scan033 & 23.45  & 7.192  & 21.79  & 22.11  & 0.911  & 0.356  & 0.902  & 0.907  & 0.075  & 0.860  & 0.081  & 0.081  \\
        scan047 & 22.91  & 7.470  & 24.08  & 24.49  & 0.909  & 0.409  & 0.905  & 0.913  & 0.104  & 0.510  & 0.094  & 0.091  \\
        scan062 & 23.71  & 25.85  & 25.74  & 25.56  & 0.907  & 0.919  & 0.914  & 0.909  & 0.076  & 0.248  & 0.066  & 0.070  \\
        scan077 & 23.40  & 13.41  & 23.11  & 21.50  & 0.928  & 0.715  & 0.920  & 0.907  & 0.100  & 0.882  & 0.101  & 0.140  \\
        scan095 & 24.00  & 17.49  & 24.51  & 20.80  & 0.867  & 0.659  & 0.873  & 0.769  & 0.153  & 0.693  & 0.143  & 0.240  \\
        scan109 & 30.13  & 28.32  & 28.86  & 28.98  & 0.911  & 0.902  & 0.899  & 0.901  & 0.151  & 0.166  & 0.162  & 0.158  \\
        scan123 & 30.64  & 17.16  & 31.58  & 30.83  & 0.910  & 0.580  & 0.910  & 0.913  & 0.212  & 0.228  & 0.198  & 0.211  \\
        \midrule
        Average & \second 25.00 \alignmark & 17.98  & \first 25.03 \alignmark & 24.57 & \first 0.901 \alignmark & 0.699 & \second 0.895 \alignmark & 0.887 & \second 0.116 \alignmark & 0.443 & \first 0.114 \alignmark & 0.128 \\
        \bottomrule
    \end{tabular}}
    \par\vspace{1mm}
    {\footnotesize\raggedright \first best, \second second best.\par}
\end{table*}

We evaluate InstantSfM against state-of-the-art SfM systems on multiple datasets, assessing both reconstruction efficiency and efficiency. We conduct comparisons of VGGSfM \cite{wang2024vggsfm}, COLMAP \cite{schonberger2016sfm}, GLOMAP \cite{pan2024global}, and our InstantSfM on a platform with an Intel Xeon Platinum 8480C and an NVIDIA H200 GPU (140.4 GB), to avoid out-of-memory issues of VGGSfM \cite{wang2024vggsfm} when handling large-scale scenes. For a fair comparison with COLMAP and GLOMAP, we use RootSIFT for feature extraction and keep the same feature matching results for the three methods. \\
$\textbf{Evaluation Criteria.}$ We evaluate on (a) rendering quality with 3DGS~\cite{ye2025gsplat}, (b) 3D map reconstruction accuracy, and (c) camera pose accuracy. We report novel view synthesis (NVS) metrics (e.g., PSNR, SSIM, and LPIPS) by training 3D Gaussian Splatting~\cite{kerbl20233d} using the estimated camera poses and 3D map reconstruction points as an initialization. We also measure the Chamfer distance on datasets with reliable ground-truth point cloud scans to evaluate 3D map reconstruction accuracy, and report pose AUC@10 when ground-truth camera poses are available.
Fig.~\ref{fig:qualitative} provides qualitative reconstruction examples across datasets.

\subsection{Performance on Various Datasets}

\noindent
\textbf{MipNeRF360}~\cite{barron2022mip} is a 360-degree object-centric dataset of 7 scenes. We use $4\times$ downsampled images following \cite{wang2024vggsfm} and report NVS metrics in Table~\ref{tab:mipnerf360_nvs_table}. This setting includes VGGSfM because all methods can be compared through the same downstream 3DGS NVS protocol; our system achieves the best average performance across the reported metrics.

\noindent\textbf{DTU}~\cite{Jensen_2014_CVPR} is an object-centric dataset of 119 scans with ground-truth geometry from structured light scanning. We evaluate under the medium lighting condition (condition 5) and report NVS metrics in Table~\ref{tab:dtu_rendering_metric}.

\begin{table}[t]
    \centering
    \caption{Chamfer Distance ($\downarrow$) comparison on ScanNet.}
    \vspace{-1mm}
    \label{tab:scannet}
    \resizebox{1.0\linewidth}{!}{
    \begin{tabular}{lcccc}
        \toprule
         & COLMAP & GLOMAP & Ours (w/o depth) & Ours (w/ depth) \\
        \midrule
        \texttt{\textbf{0001$\_$00}}  & 3.358 & \textcolor{gray}{FAILED} & 0.557 & 0.987 \\
        \texttt{\textbf{0001$\_$01}}  & \textcolor{gray}{FAILED} & \textcolor{gray}{FAILED} & 0.768 & 0.646 \\
        \texttt{\textbf{0002$\_$00}}  & \textcolor{gray}{FAILED} & \textcolor{gray}{FAILED} & 0.244 & 0.208 \\
        \texttt{\textbf{0002$\_$01}}  & \textcolor{gray}{FAILED} & \textcolor{gray}{FAILED} & 0.205 & 0.195 \\
        \texttt{\textbf{0003$\_$00}}  & 0.4139 & \textcolor{gray}{FAILED} & 2.353 & 1.098 \\ 
        \texttt{\textbf{0003$\_$01}}  & 0.2478 & \textcolor{gray}{FAILED} & 1.211 & 1.047 \\ 
        \texttt{\textbf{0003$\_$02}}  & 0.2337 & \textcolor{gray}{FAILED} & 0.556 & 0.428 \\ 
        \midrule
        \texttt{\textbf{Average}}    & \textcolor{gray}{FAILED} & \textcolor{gray}{FAILED} & \second 0.842 & \first 0.658  \\
        \bottomrule
    \end{tabular}
}
    \par\vspace{1mm}
    {\footnotesize\raggedright \first best, \second second best. \textcolor{gray}{FAILED}: optimization failure with no output. Ours (w/o depth): InstantSfM without depth prior. Ours (w/ depth): InstantSfM with depth prior.\par}
\end{table}

\begin{table}[!t]
    \centering
    \caption{Chamfer Distance ($\downarrow$) and pose AUC@10 ($\uparrow$) comparison on ScanNet++.}
    \vspace{-1mm}
    \label{tab:scannet++}
    \resizebox{1.0\linewidth}{!}{
    \begin{tabular}{lcccccccc}
        \toprule
        & \multicolumn{2}{c}{COLMAP}
        & \multicolumn{2}{c}{GLOMAP}
        & \multicolumn{2}{c}{Ours (w/o dyn. BA)}
        & \multicolumn{2}{c}{Ours (dyn. BA)} \\
        \cmidrule(lr){2-3}\cmidrule(lr){4-5}\cmidrule(lr){6-7}\cmidrule(lr){8-9}
        & CD & AUC@10 & CD & AUC@10 & CD & AUC@10 & CD & AUC@10 \\
        \midrule
        \multicolumn{9}{c}{\textit{Without GT intrinsics}} \\
        \midrule
        \texttt{\textbf{0f69ae...}}  & 0.21 & 80.60 & 0.19 & 87.01 & 0.20 & 88.95 & 0.20 & 88.87 \\
        \texttt{\textbf{ff1765...}}  & 0.56 & 21.35 & 1.58 & 71.35 & 1.85 & 24.43 & 1.75 & 25.31 \\
        \texttt{\textbf{8c3169...}}  & 1.85 & 54.96 & 1.89 & 31.06 & 0.36 & 58.44 & 0.35 & 61.79 \\
        \texttt{\textbf{00777c...}}  & 1.00 & 12.42 & 3.29 & 7.66  & 3.12 & 0.39  & 3.27 & 0.00  \\
        \texttt{\textbf{00a231...}}  & 0.31 & 70.67 & 0.35 & 73.65 & 0.40 & 68.32 & 0.39 & 68.70 \\
        \midrule
        \texttt{\textbf{Average}}    & \first 0.79 & 48.00 & 1.46 & \first 54.15 & \second 1.19 & 48.11 & \second 1.19 & \second 48.93 \\
        \midrule
        \multicolumn{9}{c}{\textit{With GT intrinsics}} \\
        \midrule
        \texttt{\textbf{0f69ae...}}  & 0.16 & 89.43 & 0.16 & 89.52 & 0.18 & 89.43 & 0.18 & 89.44 \\
        \texttt{\textbf{ff1765...}}  & 0.29 & 89.64 & 0.28 & 89.48 & 0.29 & 88.45 & 0.29 & 88.38 \\
        \texttt{\textbf{8c3169...}}  & 0.21 & 87.73 & 1.98 & 71.23 & 0.27 & 81.39 & 0.27 & 82.08 \\
        \texttt{\textbf{00777c...}}  & 0.51 & 89.13 & 0.55 & 89.09 & 0.56 & 88.99 & 0.60 & 89.02 \\
        \texttt{\textbf{00a231...}}  & 0.29 & 90.25 & 0.30 & 90.26 & 0.33 & 90.11 & 0.33 & 90.11 \\
        \midrule
        \texttt{\textbf{Average}}    & \first 0.29 & \first 89.24 & 0.65 & 85.92 & \second 0.32 & 87.67 & 0.33 & \second 87.81 \\
        \bottomrule
\end{tabular}
}
\par\vspace{1mm}
{\footnotesize\raggedright \first best, \second second best.\par}
\vspace{-15pt}
\end{table}

\noindent\textbf{ScanNet}~\cite{dai2017scannet} is a large-scale RGB-D indoor scan dataset. On ScanNet (Table~\ref{tab:scannet}), both COLMAP and GLOMAP fail on most scenes: GLOMAP due to Ceres solver divergence in its ViewGraphCalibration step, and COLMAP due to incomplete reconstructions, while InstantSfM succeeds on all scenes and improves further with depth priors.

\noindent\textbf{ScanNet++}~\cite{yeshwanthliu2023scannetpp} provides high-fidelity indoor scans with ground-truth poses registered by a laser scanner. We use this dataset to ablate dynamic parameter extraction by comparing InstantSfM with and without dynamic BA in Table~\ref{tab:scannet++}; metric depths are not used in these experiments. Here, ``dyn. BA'' denotes the proposed iterative outlier removal inside the LM loop. The feature improves AUC@10 from 48.11 to 48.93 in the unknown-intrinsics setting and from 87.67 to 87.81 in the known-intrinsics setting. It improved optimization stability under changing visibility: it consistently improves pose AUC in both intrinsic settings while preserving comparable Chamfer distance.

\subsection{Running Time Analysis}
\begin{figure}[t]
    \centering
    \begin{minipage}[t]{0.7\linewidth}
        \centering
        \includegraphics[width=\linewidth]{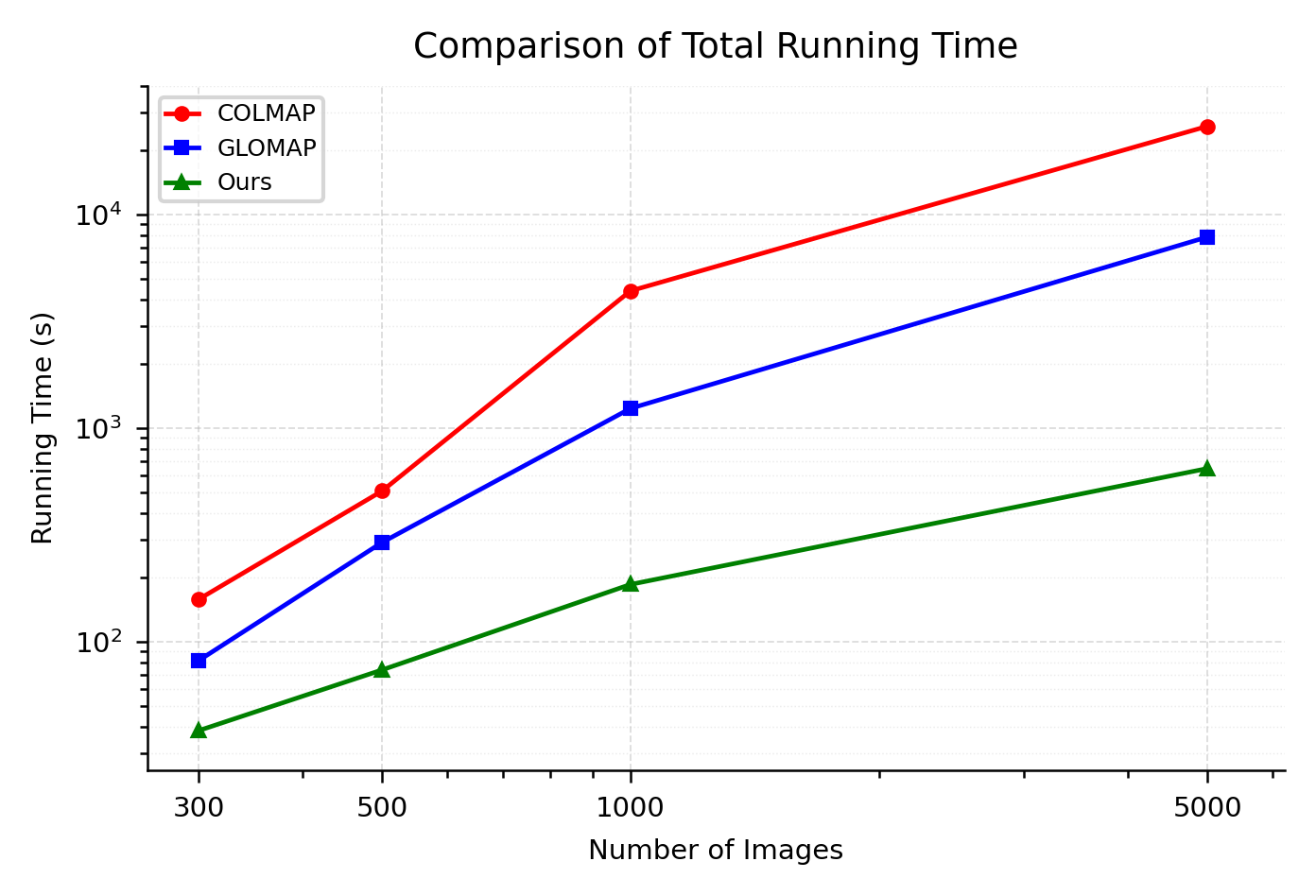} 
    \end{minipage}

    \vspace{-3mm}
    \caption{
     Comparisons of total SfM runtime of COLMAP, GLOMAP and InstantSfM on different numbers of images. Runtime is in \textbf{log space}.
    }
    \label{fig:compare_total_time}
    \vspace{-3mm}
\end{figure}

\noindent\textbf{Runtime Comparison.}
We compare InstantSfM's running time with COLMAP and GLOMAP on scenes ranging from 100 to 5,000 images, as shown in Fig.~\ref{fig:compare_total_time}. We use 1DSfM because it provides large internet photo collections with controllable image counts, which is suitable for measuring runtime scalability with scene size, following~\cite{zhan2026bundle}. \\
\textbf{Runtime Comparisons with GPU-accelerated Methods.}
We report comparisons with COLMAP and GLOMAP equipped with GPU-accelerated Ceres Solver. COLMAP and GLOMAP lag behind our method, as shown in Table~\ref{tab: comp2}. Note that 1DSfM does not provide ground-truth camera poses, so we do not report pose accuracy in this experiment.

\begin{table}[ht]
\centering
\caption{GPU runtime(s) comparison on 1DSfM~\cite{wilson2014robust}}
\resizebox{1.0\linewidth}{!}{
\begin{tabular}{lcccc}
\toprule
Scene & \# Images & COLMAP & GLOMAP & Ours  \\
\midrule
Alamo & 2915 & 12855.4 & \second 1599.5	& \first \textbf{597.0}  \\
Madrid\_Metropolis & 1344 & 1661.5 & \second 440.7 & \first \textbf{222.9}  \\
Union\_Square & 789 & 4696.7 & \second 966.4 & \first \textbf{571.3}  \\
\bottomrule
\end{tabular}
}
\par\vspace{1mm}
\vspace{-4mm}
\label{tab: comp2}
\end{table}

\section{Conclusion}
\label{sec:Conclusion}

We presented InstantSfM, a GPU-native and PyTorch-compatible global SfM system. By combining a depth-constrained Jacobian structure for metric scale recovery with dynamic parameter extraction for robust outlier handling, InstantSfM enables a complete and numerically stable global SfM within a unified GPU optimization framework.
Experiments show InstantSfM achieves competitive reconstruction quality while delivering up to ${\sim}40\times$ speedup over COLMAP and up to $12\times$ over GLOMAP on large-scale scenes.

The current system is designed for single-node execution, which may limit scalability to extremely large problems. Dynamic parameter extraction reduces the active optimization state by removing invalid cameras, points, and residuals before each linear solve, but it does not remove the fundamental memory bound of single-GPU BA. A promising future direction is improving the efficiency of the underlying optimization infrastructure, i.e., how it materializes the Jacobian, normal equations, and Schur-reduced system on a single GPU. Moreover, future versions of the optimization library are expected to further reduce this pressure through distributed optimization, which remains important future work.

\bibliographystyle{IEEEtran}
\bibliography{custom}

\end{document}